\documentclass[letterpaper, 10 pt, conference]{ieeeconf}  % Comment this line out if you need a4paper

\IEEEoverridecommandlockouts                              % This command is only needed if 
\usepackage{graphicx}
\usepackage{wrapfig}
\usepackage{eucal}
\usepackage{color}
\usepackage{amsmath}
\usepackage{lipsum} 
\usepackage[T1]{fontenc}
\usepackage{newtxtext}
\usepackage[cmintegrals]{newtxmath}
\usepackage{bm} 
\usepackage[linesnumbered,ruled,vlined]{algorithm2e}
\usepackage{subfigure}
\usepackage[utf8]{inputenc}
\usepackage{graphicx}
\usepackage{graphics}
\usepackage{siunitx}
\usepackage{multirow}
\usepackage{float}
\usepackage{caption}
\usepackage{gensymb}
\usepackage{hyperref}
\usepackage{epstopdf}
\usepackage{enumerate}
\usepackage{multicol,multirow}
\usepackage{booktabs}
\usepackage{cite}
\usepackage{graphicx}
\usepackage{textcomp}
\usepackage{xcolor}
\usepackage{wrapfig}
\usepackage{amsmath}
\usepackage{cleveref}

\usepackage{paralist}

\usepackage{graphicx}
\usepackage{wrapfig}
\usepackage{eucal}
\usepackage{color}
\usepackage{amsmath}
\usepackage{lipsum} 
\usepackage[T1]{fontenc}
\usepackage{newtxtext}
\usepackage[cmintegrals]{newtxmath}
\usepackage{bm} 
\usepackage[linesnumbered,ruled,vlined]{algorithm2e}
\usepackage{subfigure}
\usepackage[utf8]{inputenc}
\usepackage{graphicx}
\usepackage{graphics}
\usepackage{siunitx}
\usepackage{multirow}
\usepackage{float}
\usepackage{caption}
\usepackage{gensymb}
\usepackage{hyperref}
\usepackage{epstopdf}

\usepackage{multicol,multirow}
\usepackage{booktabs}
\usepackage{textcomp}
\usepackage{xcolor}
\usepackage{wrapfig}
\usepackage{amsmath}
\usepackage{cleveref}

\graphicspath{{}}

\newcommand{\vect}[1]{\boldsymbol{\mathit{\bm{#1}}}}

\definecolor{red}{rgb}{1.00,0.00,0.00}
\definecolor{lightred}{rgb}{1.00,0.3,0.3}
\definecolor{blue}{rgb}{0.00,0.00,1.00}
\definecolor{green}{rgb}{0.1,0.50,0.1}
\definecolor{yellow}{rgb}{0.5,0.5,0.0}
\definecolor{white}{rgb}{1,1,1}
\definecolor{gray}{rgb}{0.6,0.6,0.6}

\usepackage{array, makecell, rotating}
\setcellgapes{2pt}

\title{\LARGE \bf
 Shape-Aware Whole-Body Control for Continuum Robots with Application in Endoluminal Surgical Robotics}

\author{Mohammadreza Kasaei$^{1}$, Mostafa Ghobadi$^{2}$, and Mohsen~Khadem$^{1}$
\thanks{$^{1}$ Mohammadreza Kasaei and Mohsen Khadem are with the School of Informatics, University of Edinburgh, UK. Email: \{m.kasaei, mohsen.khadem\}@ed.ac.uk}%
\thanks{$^{2}$ Mostafa Ghobadi is with Playground Robotics, California, USA. Email: mostafa\_ghobadi@hotmail.com}
 \thanks{
 This work is supported by the Medical Research Council [MR/T023252/1]}
}

\newenvironment{customindent}[1]%
  {\begin{list}{}%
     {\setlength{\leftmargin}{#1}}%
     \item[]%
  }
  {\end{list}}

\begin{document}

\maketitle
\thispagestyle{empty}
\pagestyle{empty}

%%%%%%%%%%%%%%%%%%%%%%%%%%%%%%%%%%%%%%%%%%%%%%%%%%%%%%%%%%%%%%%%%%%%%%%%%%%%%%%%
\begin{abstract}
This paper presents a shape-aware whole-body control framework for tendon-driven continuum robots with direct application to endoluminal surgical navigation. Endoluminal procedures, such as bronchoscopy, demand precise and safe navigation through tortuous, patient-specific anatomy where conventional tip-only control often leads to wall contact, tissue trauma, or failure to reach distal targets. To address these challenges, our approach combines a physics-informed backbone model with residual learning through an Augmented Neural ODE, enabling accurate shape estimation and efficient Jacobian computation. A sampling-based Model Predictive Path Integral (MPPI) controller leverages this representation to jointly optimize tip tracking, backbone conformance, and obstacle avoidance under actuation constraints. A task manager further enhances adaptability by allowing real-time adjustment of objectives, such as wall clearance or direct advancement, during tele-operation. Extensive simulation studies demonstrate millimeter-level accuracy across diverse scenarios, including trajectory tracking, dynamic obstacle avoidance, and shape-constrained reaching. Real-robot experiments on a bronchoscopy phantom validate the framework, showing improved lumen-following accuracy, reduced wall contacts, and enhanced adaptability compared to joystick-only navigation and existing baselines. These results highlight the potential of the proposed framework to increase safety, reliability, and operator efficiency in minimally invasive endoluminal surgery, with broader applicability to other confined and safety-critical environments.
\end{abstract}

%%%%%%%%%%%%%%%%%%%%%%%%%%%%%%%%%%%%%%%%%%%%%%%%%%%%%%%%%%%%%%%%%%%%%%%%%%%%%%%%
\section{Introduction}
\label{sec:intro}

Continuum robots made from compliant materials can conform to tortuous anatomy while tolerating contact forces~\cite{laschi2016soft,rus2015design}. These properties make them particularly attractive for endoluminal surgery, where safe navigation through delicate, patient-specific pathways is essential~\cite{sharma2023novel,10335932,mklung}. Similar advantages have driven their use in search and rescue~\cite{yamauchi2022development} and exploration of hazardous environments~\cite{wooten2018exploration}. However, the same compliance that enables adaptability also poses major challenges: continuum robot shape is governed by nonlinear, high-dimensional mechanics that are difficult to model and control, limiting reliable whole-body coordination in clinical and other constrained settings~\cite{george2018control,review2,mengaldo:hal-03921606}. Therefore, achieving reliable shape control is challenging. 

\subsection{Shape Estimation}
Recent research on continuum robot shape estimation has followed two main paradigms: physics-based modeling and data-driven learning, with hybrid strategies increasingly explored at their intersection.

Model-based methods rely on analytical kinematic and dynamic formulations. Piecewise constant curvature (PCC) models~\cite{webster2010design,katzschmann2019dynamic} are widely used for their simplicity but degrade in accuracy under external interaction. Cosserat rod models~\cite{till2019real,rucker2,mohsen1} capture mechanics more faithfully, though at higher computational cost and with demanding calibration. Closed-loop controllers with shape sensing (e.g., fiber-Bragg gratings or vision) improve robustness but add hardware complexity. Recent advances focus on adaptivity: Zhai \emph{et al.} applied online Jacobian error compensation via Kalman filtering~\cite{Zhai2025}, Wang \emph{et al.} introduced a hybrid adaptive controller for tool–tissue interaction~\cite{Wang2021hybrid}, and Alora and Pavone proposed spectral submanifold reduction for real-time model predictive control~\cite{Alora2025}. These works illustrate a shift toward balancing fidelity with tractability.
\begin{figure}[!t]
    \centering
    \includegraphics[width=\linewidth]{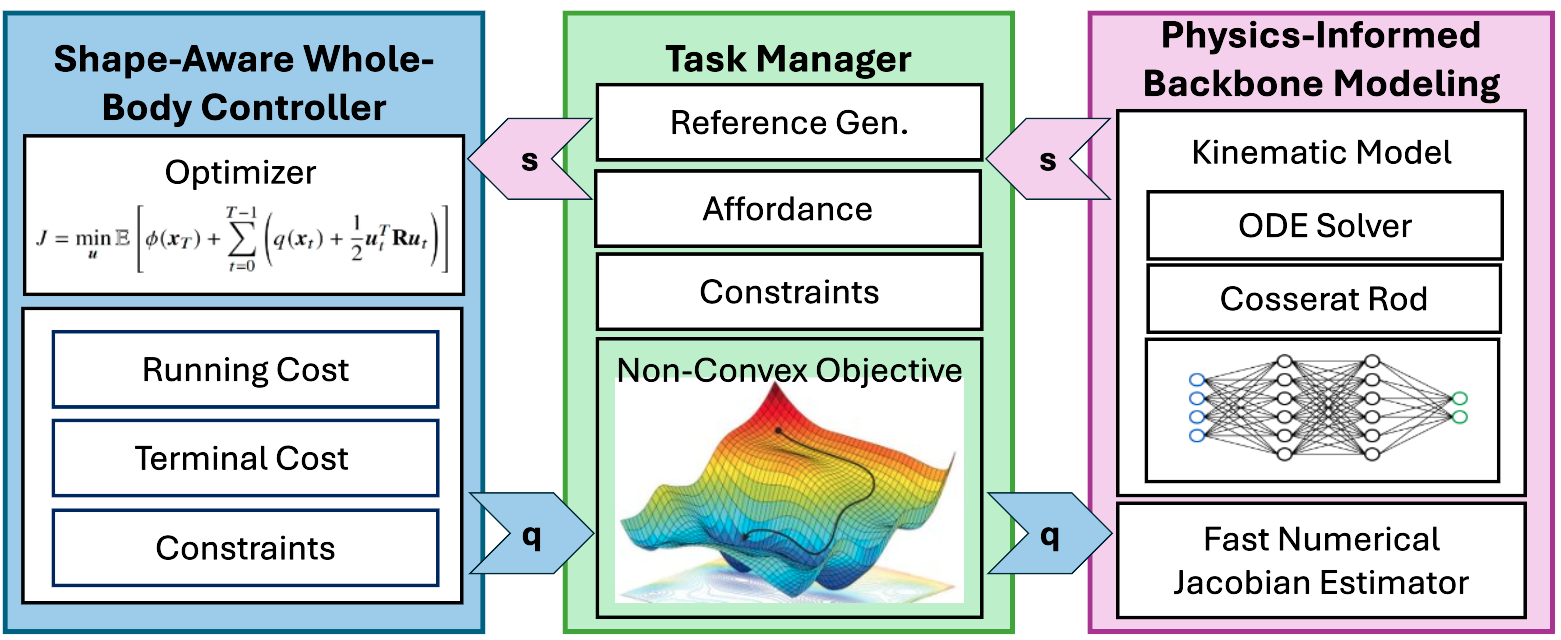}
\caption{Overall architecture of the proposed framework, consisting of three main modules. 
The backbone modeling combines physics-based Cosserat modeling with neural residuals to provide accurate backbone reconstruction. 
The Whole-Body Controller employs MPPI to jointly optimize tip tracking, shape regulation, and obstacle avoidance under actuation constraints. 
The Task Manager interfaces with the operator, enabling online adjustment of control objectives such as clearance, compliance, or direct advancement.}

    \label{fig:overview}
    \vspace{-3mm}
\end{figure}

Learning-based approaches avoid explicit modeling by deriving surrogate models or controllers from data. Reinforcement learning (RL) has shown promise, from early demonstrations of tip positioning in simulation~\cite{Satheeshbabu2019} to more recent controllers achieving sub-millimeter accuracy~\cite{Wei2023}. Ensemble and fuzzy RL methods further improve robustness~\cite{Morimoto2021}, but purely data-driven control remains limited by large training requirements, poor generalization, and lack of safety guarantees—critical issues in surgical settings.

Hybrid and data-driven modeling approaches aim to overcome these limitations. Maghooli \emph{et al.} combined deep RL with a Jacobian-based controller, achieving superior performance in animal-tissue experiments~\cite{Maghooli2025}, while Yu and Tan used Koopman operators to learn linear embeddings of continuum dynamics for provably robust control~\cite{Yu2025}. Other work has emphasized end-to-end data-driven estimators or physics-assisted learning~\cite{gao2024sim,habich2025generalizable}, which improve robustness to fabrication errors but still face challenges of data efficiency, generalization, and interpretability~\cite{kasaei2023data,kasaei2023CORL}. Collectively, these efforts point to hybrid strategies that combine the interpretability and safety of physics-based models with the adaptability of learning.

\subsection{Control}
The control of continuum robots is complicated by nonlinear kinematics, high-dimensional dynamics, and strong input–deformation coupling. Classical formulations such as PCC and Cosserat models provide useful abstractions, but struggle to capture full-body nonlinearities and often incur high computational costs. To overcome these limitations, reinforcement learning and other data-driven strategies have been proposed to directly learn control policies from data~\cite{feizi2025deep,balint1,balint2,bern2020soft,thuruthel2017learning,bruder2019nonlinear,morimoto2021model,buchler2022learning}. While promising, these methods typically emphasize tip tracking, require large-scale training, and lack flexibility once deployed, making them less suitable for tasks where whole-body shape plays a critical role.

Model Predictive Path Integral (MPPI) control has emerged as a powerful alternative for continuum robots. As a sampling-based variant of MPC, MPPI evaluates thousands of candidate trajectories in parallel, making it naturally suited to their nonlinear, high-dimensional dynamics. Unlike learning-based methods, it requires no offline training, and its explicit cost function provides transparency and interpretability. A distinctive advantage of MPPI is its capacity to reweight objectives on the fly, allowing the controller to adapt priorities—such as tip accuracy, shape conformity, or obstacle avoidance—according to task demands. This adaptability is especially valuable in tele-operated procedures, where operator intent may change during execution. By moving beyond purely tip-centric strategies and supporting real-time rebalancing of objectives, MPPI offers a practical and flexible framework for shape-aware continuum robot control in uncertain environments.

\subsection{Contributions}  
Motivated by applications such as robotic endoluminal surgery, where robots must operate within highly constrained anatomical pathways, we advance continuum robot control beyond tip-centric approaches by introducing a unified framework that integrates physics-informed backbone modeling, whole-body control, task-level adaptability, and shared-control tele-operation. While the focus is on medical navigation, the framework is broadly applicable to inspection in confined environments and search-and-rescue operations. The main contributions are:

\begin{customindent}{-3pt}
\begin{itemize}
    \item \textbf{Physics-informed backbone modeling:}  
    We represent the continuum backbone with a segment-wise ODE formulation that integrates curvature and orientation frames to capture full-body deformation. This structured model can be augmented with learnable residuals to account for unmodeled dynamics, yielding a representation that is both interpretable and adaptable. Unlike tip-only models, it provides access to the entire backbone configuration, which is critical for whole-body control.  
    \item \textbf{Shape-aware whole-body control:}  
    A Model Predictive Path Integral (MPPI) controller that simultaneously optimizes tip tracking, backbone regulation, and obstacle avoidance, while respecting actuation constraints.  
    \item \textbf{Task manager for online adaptation:}  
    A high-level supervisory module that adjusts objective weights in real time, enabling operators to flexibly prioritize compliance, directional bias, or selective segment freedom in tele-operated settings.  
    \item \textbf{Open-source framework:}  
    A publicly available implementation that provides a reproducible platform to accelerate future research on continuum robot modeling and control.  
\end{itemize}
\end{customindent}
The structure of the paper is as follows. Section~\ref{sec:method} introduces the proposed methodology, outlining the development of the physics-informed backbone modeling and whole-body control. Section~\ref{sec:sim} reports simulation studies that examine the framework’s performance across multiple tasks. The experimental setup and corresponding real-robot results are presented in Section~\ref{sec:exp}. A comparative analysis against existing approaches is provided in Section~\ref{sec:comp}, and concluding remarks are drawn in Section~\ref{sec:conclusion}.

\section{Methodology}
\label{sec:method}
This section explains detail of our proposed methodology. The overall architecture of the proposed framework is shown in Figure~\ref{fig:overview}. As illustrated, it is composed of three main components: the \textit{Physics-informed Backbone Modeling}, the \textit{Whole-Body Controller}, and the \textit{Task Manager}. The following subsections describe each component in detail.

\subsection{Physics-informed backbone modeling}
The kinematics of continuum robots are commonly described by the \textit{Cosserat rod model}, which represents the backbone as a continuously deformable curve. At each point along the centerline, a moving frame is attached such that its local $z$-axis remains tangent to the curve. The pose of this frame is expressed by a homogeneous transformation $\mathbf{T}(s,t) \in \text{SE}(3)$:
\begin{equation}
\mathbf{T}(s,t) =
\begin{bmatrix}
\mathbf{R}(s,t) & \vect{p}(s,t) \\
\mathbf{0}_{1 \times 3} & 1
\end{bmatrix},
\end{equation}
where $\vect{p}(s,t) \in \mathbb{R}^3$ denotes the position and $\mathbf{R}(s,t) \in \mathit{SO}(3)$ the orientation. Here, $s \in [0,\ell]$ is the arc-length parameter, and $\ell$ is the total backbone length. The frame evolves according to the Cosserat rod equations~\cite{rucker2}:
\begin{equation}
\mathbf{T}'(s,t) = \mathbf{T}(s,t)[\xi(s,t)]_\times, 
\quad
\vect{u}'(s,t) = h(s,\vect{u},q(t)),
\end{equation}
where $\xi(s,t) = (\vect{u}(s,t), e_3)$ is the body twist, $[\cdot]_\times$ is the $4\times 4$ twist matrix~\cite{mrbook}, and $\vect{u}(t) = [u_x(t), u_y(t), 0]^T$ encodes curvature in the $x$--$y$ plane. The nonlinear mapping $h$ relates curvature changes to arc-length, curvature state, and actuation inputs. The configuration at any point $S$ is obtained by integration, with the tip position given by
\begin{equation}
\vect{x}(t) = \vect{p}(\ell,t) = \int_{0}^{\ell} \vect{p}(s,t)\,ds.
\label{eq:math_model}%
\end{equation}
For multi-segment robots, each segment has its own kinematic parameters, and continuity is ensured by propagating the terminal state of one segment as the initial condition of the next.  

\medskip
\noindent
\textbf{Learning Model Mismatch.}
While the Cosserat rod formulation provides a rigorous description of backbone kinematics, discrepancies inevitably arise due to unmodeled effects such as material nonlinearities, external loads, friction, or actuation imperfections. To address this issue, we introduce an Augmented Neural ODE (ANODE)~\cite{dupont2019augmented} framework, which retains the analytical model as a baseline but employs a neural network to learn residual dynamics and compensate for mismatch.  

Let the nominal backbone evolution be governed by
\begin{equation}
{\vect{p}}^{'}(s,t) = f(\vect{p}, \vect{u}(t)), 
\quad f: \mathbb{R}^3 \times \mathbb{R}^3 \to \mathbb{R}^3,
\end{equation}
with initial conditions $\vect{p}_0, \vect{u}_0$. Since $f$ cannot fully capture the real behavior, we augment it with a neural correction term $f_\theta$, parameterized as a time-dependent MLP:
\begin{equation}
\frac{\partial \vect{p}(t)}{\partial t} = f(\vect{p}(t), \vect{u}(t)) + f_{\theta}(\vect{p}(t), \vect{u}(t), t).
\end{equation}

Backbone reconstruction is then carried out by numerical integration along the arc length. For a segment $[S_i, S_{i+1}]$:
\begin{equation}
\hat{\vect{p}}_{S_{i+1}} = \operatorname{ODESolver}(f+f_{\theta}, (\vect{p}_{S_i}, \vect{u}_{S_i}), (S_i, S_{i+1})).
\label{eq:ode_solver}
\end{equation}
The correction term ensures that deviations between the nominal Cosserat prediction and the actual backbone evolution are adaptively reduced, improving overall accuracy.

\medskip
\noindent
\textbf{Training.}
The $f_\theta$ is trained to minimize the mismatch between predicted and measured backbone configurations. Specifically, the predicted states obtained from the ODE solver are compared against ground-truth backbone points distributed along the arc length:
\begin{equation}
\mathcal{L}(\theta) = \sum_{k=1}^{K} \left\| \hat{\vect{p}}(S_k) - \vect{p}(S_k) \right\|,
\end{equation}
where $K$ denotes the number of discretization points. By distributing the loss along the entire backbone rather than only at the tip, the model learns to correct residual errors across the full deformation profile.  

Equation~(\ref{eq:ode_solver}) enables batch-wise calculation across multiple samples \((\vect{u}^N = \{\vect{u}_0, \vect{u}_1, \ldots, \vect{u}_N\})\). To ensure consistent integration, all trajectories are solved over a uniform arc-length. Since the effective backbone length $\ell(t)$ may vary across samples, the maximum length in the batch is first identified. Integration is then performed for all samples up to this maximum length. For robots with shorter effective lengths, the solution is masked beyond their true length by propagating the last valid state, thereby preserving batch consistency. This strategy not only stabilizes the training process but also allows the network $f_\theta$ to learn residual dynamics reliably across heterogeneous trajectories. In addition, the batch formulation is exploited for efficient Jacobian computation using central differences, which significantly accelerates gradient evaluation.

\begin{figure*}[!t]
    \centering
    \subfigure[]{
        \includegraphics[width=0.18\textwidth]{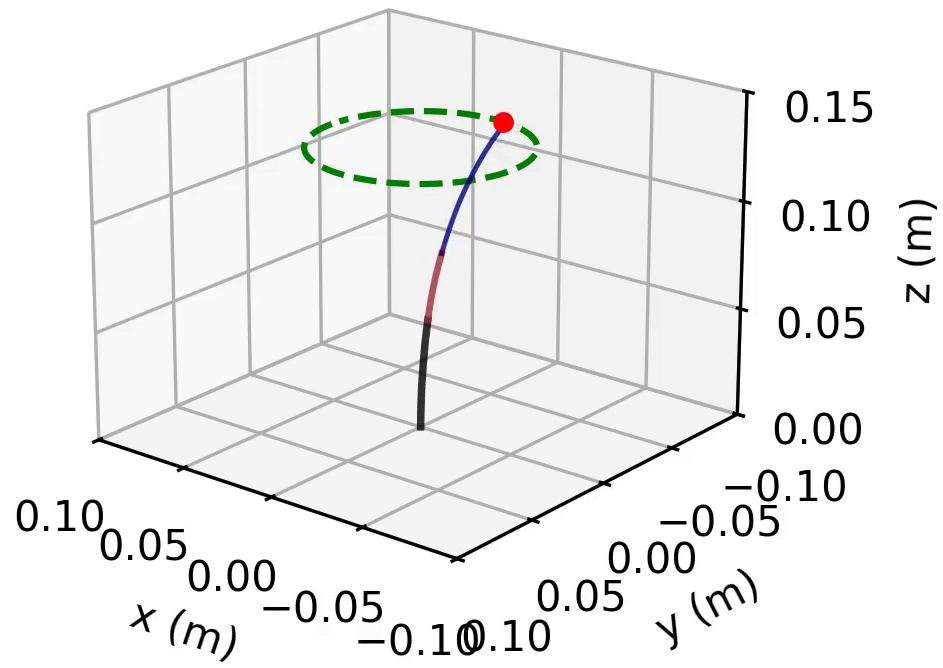}
    }
    \subfigure[]{
        \includegraphics[width=0.18\textwidth]{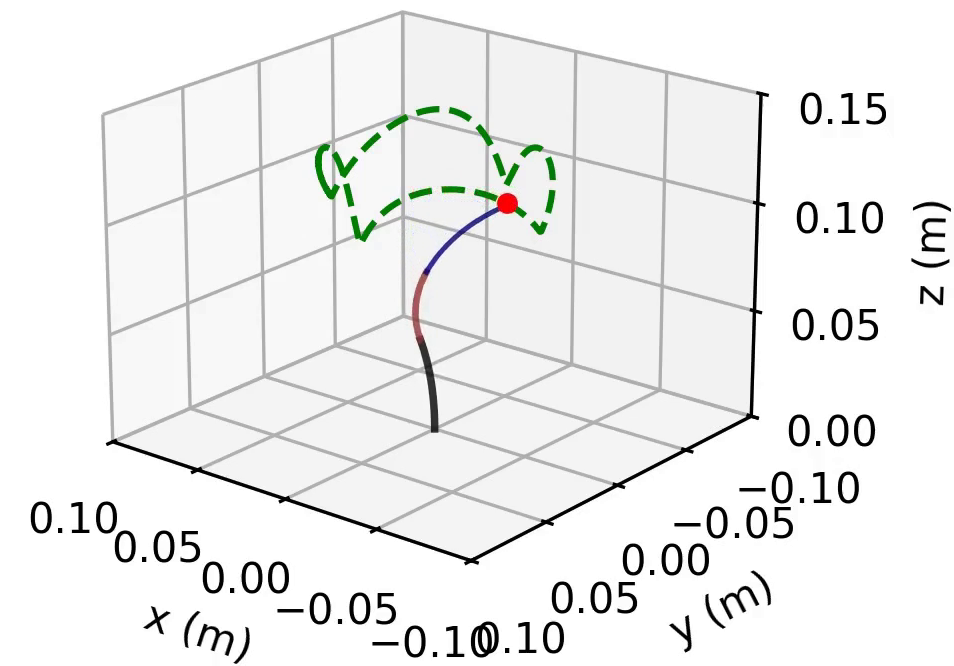}
    }
    \subfigure[]{
        \includegraphics[width=0.18\textwidth]{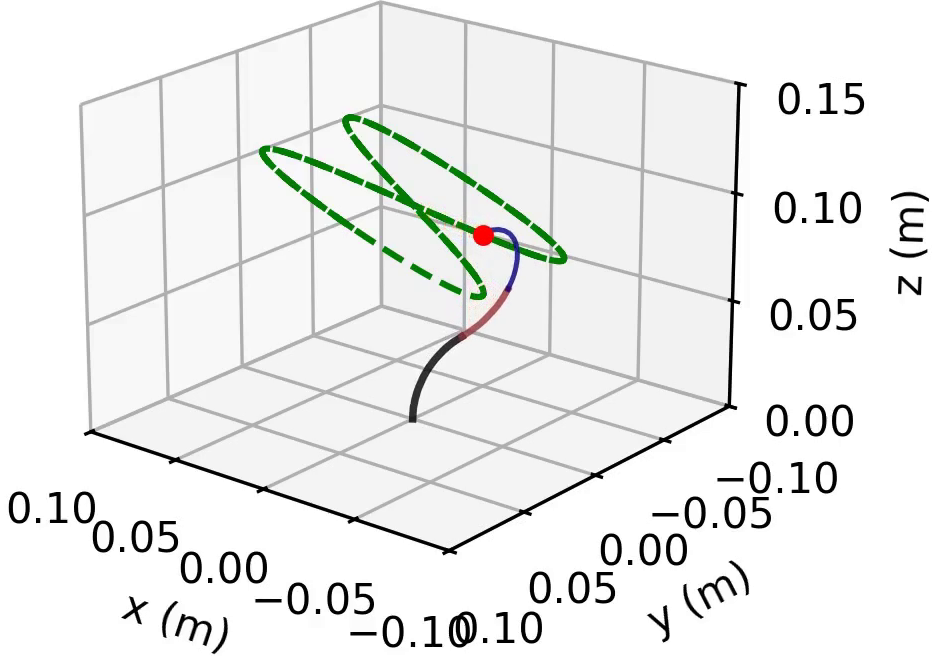}
    } 
    \subfigure[]{
        \includegraphics[width=0.18\textwidth]{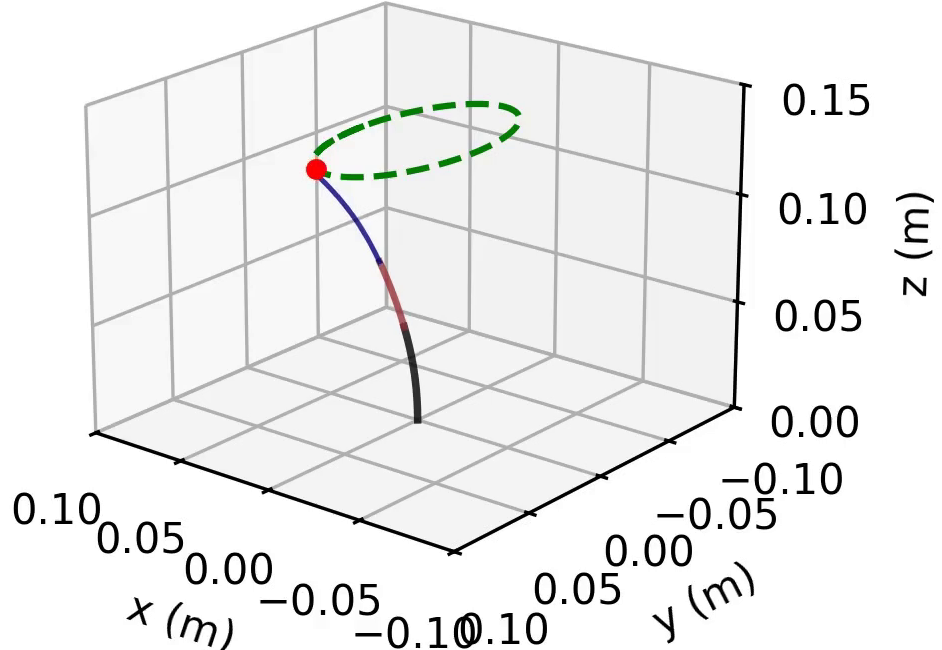}
    }
    \subfigure[]{
        \includegraphics[width=0.18\textwidth]{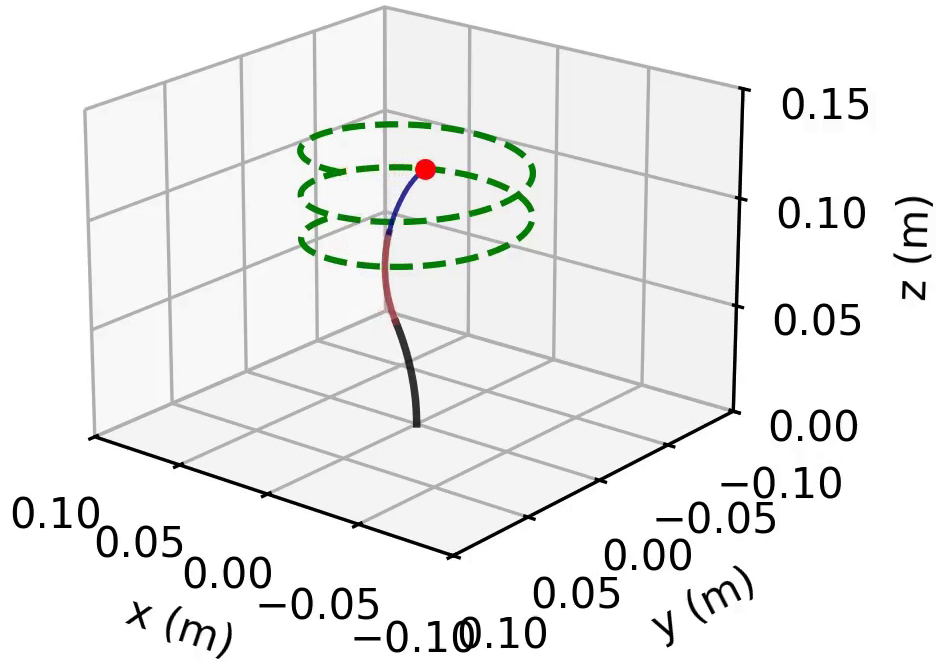}
    }
    \vspace{-2mm}
    \caption{Representative results for the trajectory tracking scenario: red markers show the current tip position, green dashed lines are the reference trajectories, and solid lines indicate the robot's body shape: (a)~circular, (b)~curvy-
edge circular, (c)~butterfly, (d)~elliptical, and (e)~helical trajectories.}
    \vspace{-5mm}
    \label{fig:traj_tracking}    
\end{figure*}

\subsection{Whole-Body Control}
To control the continuum robot motion, we formulate a shape-aware whole-body control problem where the objective is to drive the tip towards a desired reference while regulating the global backbone deformation. The robot state is represented by the downsampled backbone configuration $\mathbf{x} \in \mathbb{R}^{n_x}$, obtained by concatenating the coordinates of $m$ equidistant points along the backbone, while the control input is the actuation vector $\mathbf{q} \in \mathbb{R}^{n_u}$. The dynamics are approximated by a first-order model derived from the Jacobian,  
\begin{equation}
\dot{\mathbf{x}} = J(\mathbf{q})\,\dot{\mathbf{q}}, 
\label{eq:dynamics}
\end{equation}
where $J(\mathbf{q}) \in \mathbb{R}^{n_x \times n_u}$ is estimated online using batch central differences of~(\ref{eq:ode_solver}). This formulation provides a fast and efficient approximation of how incremental changes in actuation propagate to global backbone deformations.

The control objective over a finite horizon $T$ is defined by the cost functional
\begin{equation}
\begin{aligned}
\mathcal{J} = \sum_{t=0}^{T-1}  
& w_{\mathrm{tip}} \|\mathbf{x}^{\mathrm{tip}}_t - \mathbf{x}^{\mathrm{ref}}_t\|^2 +
w_{\mathrm{shape}} \|\mathbf{x}_t - \mathbf{x}^{\mathrm{ref}}_t\|^2 + \\
& w_{u} \|\mathbf{q}_t - \mathbf{q}_{t-1}\|^2 
+ w_{\mathrm{term}} \|\mathbf{x}^{\mathrm{tip}}_T - \mathbf{x}^{\mathrm{ref}}_T\|^2 , 
\end{aligned}
\label{eq:objective}
\end{equation}
where $\mathbf{x}^{\mathrm{tip}}_t \in \mathbb{R}^3$ denotes the tip position extracted from $\mathbf{x}_t$, $\mathbf{x}^{\mathrm{ref}}_t$ is the reference trajectory, and the weights $w_{\mathrm{tip}}, w_{\mathrm{shape}}, w_u, w_{\mathrm{term}}$ trade off tracking accuracy, shape regularity, control smoothness, and terminal performance. In addition, obstacle avoidance can be incorporated by penalizing proximity between backbone points and predefined obstacle surfaces,  
\begin{equation}
C_{\mathrm{obs}}(\mathbf{x}_t) = \sum_{i=1}^m \phi(d(\mathbf{x}_{t,i}, \mathcal{O})),
\label{eq:constraint}
\end{equation}
where $d(\cdot,\mathcal{O})$ computes the signed distance of a backbone point $\mathbf{x}_{t,i}$ to the obstacle set $\mathcal{O}$ and $\phi(\cdot)$ is a barrier-like penalty function. The optimal control input is then obtained as a weighted average over the sampled candidates. This formulation ensures that the control policy is sensitive to both tip tracking and full-body deformation, while efficiently exploiting batch Jacobian computations for fast online performance.

The resulting control problem is solved using the Model Predictive Path Integral~(MPPI) framework, which generates $K$ candidate control sequences $\{\mathbf{q}_{0:T-1}^k\}_{k=1}^K$, evaluates their predicted trajectories via~\eqref{eq:dynamics}, and assigns them importance weights according to the exponential transformation of~\eqref{eq:objective} and~\eqref{eq:constraint}. 
The control inputs are subject to physical constraints that reflect actuation limits, ensuring feasible tendon displacements and avoiding damage to the structure. We denote the admissible input set as
\begin{equation}
\mathcal{U} = \{\mathbf{q} \in \mathbb{R}^{n_u} \;|\; \mathbf{q}_{\min} \leq \mathbf{q} \leq \mathbf{q}_{\max}\},
\label{eq:input_constraints}
\end{equation}
where $\mathbf{q}_{\min}$ and $\mathbf{q}_{\max}$ are vectors of lower and upper bounds, respectively. In practice, these bounds correspond to maximum allowable tendon length variations, and in our implementation are enforced by clamping inputs. Within the MPPI framework, candidate trajectories $\mathbf{q}_{0:T-1}$ are sampled only from $\mathcal{U}$, guaranteeing that the resulting commands are physically realizable. The key hyperparameters of the model $f_\theta$ and the control settings for MPPI are summarized in Table~\ref{tab:hyper_mppi}.

\subsection{Task Manager}
While the shape-aware whole-body controller optimizes inputs to track references and regulate deformation, many tasks require dynamic adjustment of objectives according to user intent or contextual constraints. To this end, we introduce a task manager module that provides a flexible interface between the user and the controller by tuning the objective weights and constraints in real time. Specifically, the task manager exposes a set of adjustable parameters that modulate the relative importance of different terms in the cost function~\eqref{eq:objective}--\eqref{eq:constraint}.   

This design enables affordance-based behavior: for example, when the user specifies a preferred direction of motion, the task manager increases the weight of objectives aligned with that direction while suppressing motions along undesired axes. Similarly, the module can relax penalties on a particular segment to allow greater freedom of movement, while simultaneously stiffening others to ensure stability or maintain contact constraints. In practice, the adjustable parameters are implemented as a high-level policy layer that maps user commands or task descriptors into a configuration of weights $\{w_{\mathrm{tip}}, w_{\mathrm{shape}}, w_u, w_{\mathrm{term}}\}$ and constraint thresholds $\{\mathbf{q}_{\min}, \mathbf{q}_{\max}\}$.  

By separating low-level control from task-level modulation, the task manager provides an intuitive mechanism for tailoring robot behavior to specific application requirements. This allows users to guide the continuum robot not only toward spatial goals but also toward functional behaviors such as compliance, selectivity in segment motion, or directional affordances, thereby expanding the versatility of the shared-control framework.

\begin{table}[!t]
\vspace{-2mm}
\caption{Model Hyperparameters for $f_\theta$ and MPPI Control Parameters}
\centering
\resizebox{0.48\textwidth}{!}{
\begin{tabular}{|c|c|}
\hline
\multicolumn{2}{|c|}{\textbf{$f_\theta$ (ANODE)}} \\ \hline
architecture & 21@256@256@21* \\ \hline
activation & LeakyReLU(@); Tanh(*) \\ \hline
optimizer & Adam \\ \hline
learning rate & $1\times10^{-3}$ \\ \hline
batch size & 256 \\ \hline
total iterations & 20k \\ \hline
ODE solver & fixed-adams \\ \hline
\end{tabular}
\quad
\begin{tabular}{|c|c|}
\hline
\multicolumn{2}{|c|}{\textbf{MPPI Parameters}} \\ \hline
$\mu_{\text{noise}}$ & $[0,0,0]$ \\ \hline
$\Sigma_{\text{noise}}$ & $5\times10^{-5} I$ \\ \hline
$u_{\min}$ & $[-0.03, -0.015, -0.015, \dots]$ \\ \hline
$u_{\max}$ & $[+0.03, +0.015, +0.015, \dots]$ \\ \hline
$\lambda$ & 0.001 \\ \hline
horizon & 10 \\ \hline
num samples & 1000 \\ \hline
\end{tabular}}
\label{tab:hyper_mppi}
\vspace{-0mm}
\end{table}

\section{Simulations}
\label{sec:sim}
We conduct a comprehensive set of simulation studies to evaluate the performance of the proposed framework across diverse scenarios. The initial studies focus on fundamental capabilities, namely trajectory tracking and dynamic obstacle avoidance, while subsequent simulations address more advanced tasks that emphasize shape-aware whole-body control. In this way, the framework is systematically assessed in terms of its ability to handle increasingly complex interactions and constraints.

The robot used in our simulations is a cable-driven continuum manipulator modeled by~(\ref{eq:math_model}). It is composed of three serially connected segments, each with an initial length of $l_0 = 50\,\mathrm{mm}$ and a cable offset of $d = 7.5\,\mathrm{mm}$. Actuation is achieved through differential length changes of the driving cables, which generate curvature in the $x$- and $y$-directions. The continuous deformation of the manipulator is captured by integrating the kinematics of each segment along its backbone using an ODE formulation with a spatial step size of $\Delta s = 0.001$. This model provides access to both the tip position and the full body configuration, thereby enabling whole-body, shape-aware control.

\begin{figure*}[!t]
    \centering
    
        \includegraphics[width=0.95\textwidth]{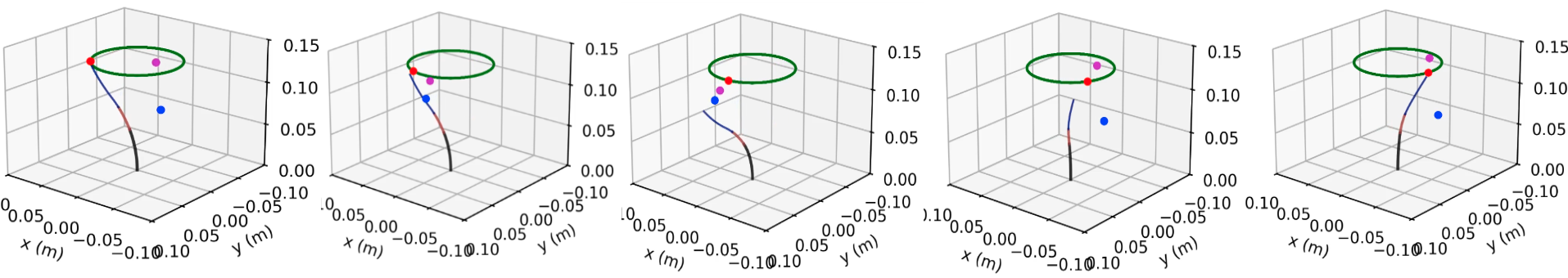}
    \caption{A representative simulation result of \emph{Dynamic obstacle avoidance}. The green curve represents the reference trajectory, and the red marker denotes the current tip position. Blue and magenta markers indicate obstacles. The solid curve illustrates the three connected body segments of the robot, highlighting how the framework coordinates whole-body motion to remain close to the desired trajectory while maintaining safe clearance from obstacles.}
    \label{fig:obs}
     \vspace{-5mm}

\end{figure*}

\begin{figure}[!t]
    \centering
    \includegraphics[width=0.95\linewidth]{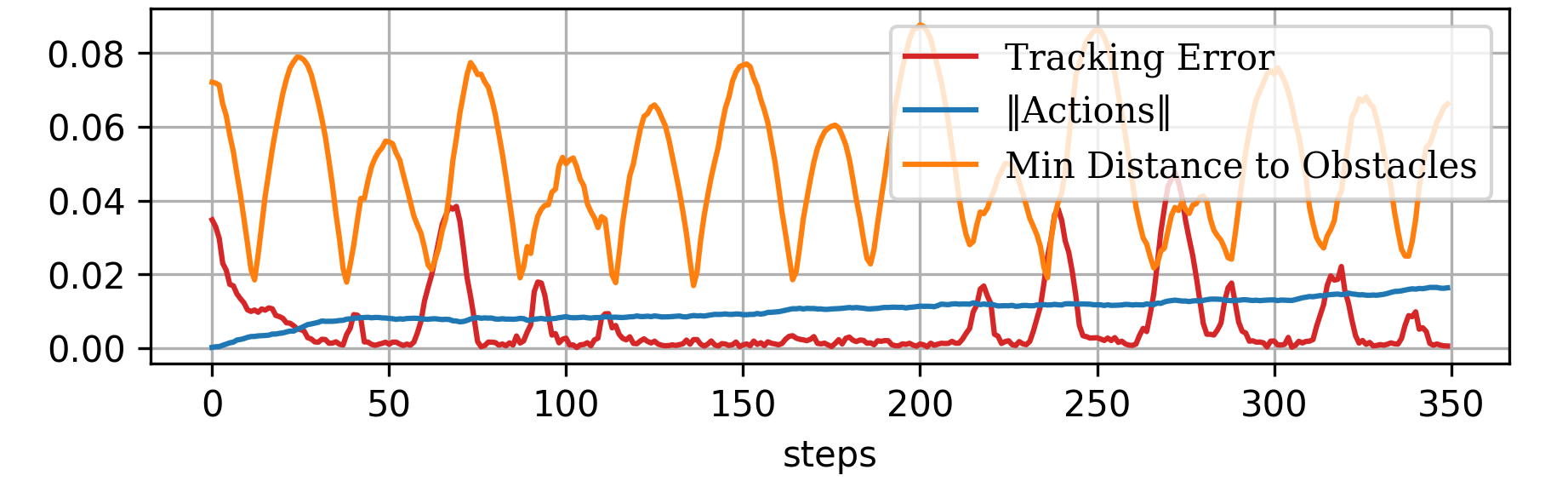}
    \caption{Performance of the dynamic obstacle avoidance task. The controller maintains millimeter-level tracking accuracy with error spikes during avoidance maneuvers. The control effort gradually increases as the robot reshapes its body to ensure clearance, while the minimum distance to obstacles consistently remains above the 0.02~m safety threshold, demonstrating successful collision avoidance.}
        \label{fig:obs_plot}
\end{figure}

\begin{customindent}{-3pt}
\begin{itemize}
     \item \textbf{Trajectory tracking:} For the trajectory tracking scenario, the robot is commanded to follow a set of representative paths in three-dimensional space. The circular path is given by 
$(r \cos(\omega t), \, r \sin(\omega t), \, 0.13)$, 
while a curvy-edge circular path extends this motion with small oscillations in $y$ and $z$, expressed as 
$(r \cos(\omega t), \, r \sin(\omega t) + \epsilon \cos(3 \omega t), \, 0.13 + c |\sin(2 \omega t)|)$. 
An elliptical path is described by 
$(a \cos(\omega t), \, b \sin(\omega t), \, 0.13 + c \sin(0.5 \omega t))$, 
capturing anisotropic oscillations along the main axes. A butterfly-shaped trajectory combines sinusoidal motions in all three directions, 
$(a \sin(2\pi t/T), \, b \cos(\pi t/T), \, 0.1 + c \sin(2\pi t/T))$. 
Finally, the helical trajectory is given by 
$(r \sin(\omega t), \, r \cos(\omega t), \, -\alpha t)$, 
representing circular motion in the $xy$-plane combined with a uniform descent along $z$.  

    \item \textbf{Dynamic obstacle avoidance:} In the dynamic obstacle avoidance task, the robot follows a circular path while avoiding collisions with two moving obstacles. Unlike tip-based tracking, the robot must adapt its entire body configuration to remain safe, ensuring that no part of the structure comes into contact with the obstacles while still maintaining the desired trajectory.

    \item \textbf{Shape-constrained reaching:} In this task, the robot is required to reach a target position while simultaneously conforming its body to a desired shape. The main difficulty of this task is that the robot must coordinate its motion to satisfy both end-point accuracy and whole-body shape constraints, ensuring that the final configuration matches the specified target.  
    \item \textbf{Tele-operated shape tracking:} In this task, the reference is generated by a virtual robot whose shape is directly controlled by the operator through a keyboard interface. The physical robot must track the target position and body configuration of this virtual counterpart. Importantly, the controller has no access to the operator’s commands; instead, it relies solely on the observed shape of the virtual robot to guide the motion. This setup enables a form of shared control, where the operator specifies the desired shape through the virtual robot and the controller ensures stable and accurate execution on the physical system.

\end{itemize}
\end{customindent}

\begin{figure*}[!t]
    \centering
    
        \includegraphics[width=0.95\textwidth]{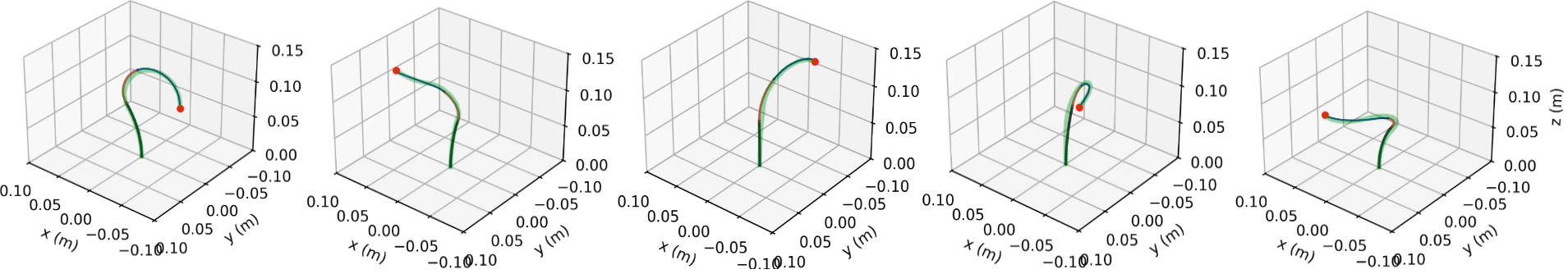}
    \caption{Simulation results of the \emph{Shape-constrained reaching} task. 
    The transparent green lines represent the reference body shapes, while the red markers indicate the target tip positions. 
    The robot adapts its configuration to simultaneously match the desired tip location and overall shape.}
    \label{fig:shape_constrained_reaching}
     \vspace{-6mm}

\end{figure*}

\subsection{Results and Discussions}

In the \textit{Trajectory tracking} scenario, we randomized the path parameters for each trajectory~(\(a,b \in [0.02,0.07]\)~m and \(c \in [0.01,0.02]\)~m) and conducted 50~trials per shape to evaluate robustness under varying curve profiles. The results, summarized in Table~\ref{tab:trajectory_tracking_sim}, show consistently accurate and repeatable tracking performance across all trajectories. The mean errors remain within 2--3~mm, the RMSE is typically in the 5--8~mm range, and the standard deviations are small, indicating robustness against trajectory variability. Among the tested paths, the helix yields the lowest error (\(\text{RMSE}=0.0052 \pm 0.0007\)~m, maximum error \(0.0422 \pm 0.0093\)~m), which can be attributed to its smooth and continuous curvature. In contrast, the butterfly trajectory proves more challenging (\(\text{RMSE}=0.0078 \pm 0.0005\)~m, maximum error \(0.0601 \pm 0.0054\)~m), reflecting the impact of abrupt curvature changes and inflection points. The circle and ellipse trajectories achieve comparable accuracy (\(\text{RMSE}=0.0056 \pm 0.0011\)~m and \(0.0059 \pm 0.0013\)~m, respectively), while the curvy-edge shape exhibits slightly higher error than the smooth closed curves due to its locally sharper turns. These results confirm that the controller achieves millimeter-level tracking precision across randomized trials, with performance degradation primarily occurring in trajectories that involve rapid curvature variations. Figure~\ref{fig:traj_tracking} shows a set of representative snapshots of this scenario.

\begin{table}[!t]
\caption{Simulation Results for Trajectory Tracking}
\centering
\renewcommand{\arraystretch}{1.1}
\setlength{\tabcolsep}{4pt}
\begin{tabular}{|c|c|c|c|}
\hline
\textbf{Trajectory} & \textbf{RMSE [m]} & \textbf{Mean Err. [m]} & \textbf{Max Err. [m]} \\ \hline
Circle     & 0.0056 $\pm$ 0.0011 & 0.0023 $\pm$ 0.0003 & 0.0488 $\pm$ 0.0104 \\ \hline
Curvy-edge & 0.0061 $\pm$ 0.0015                  & 0.0032 $\pm$ 0.0006                  & 0.0489 $\pm$ 0.0136                  \\ \hline
Butterfly  & 0.0078 $\pm$ 0.0005 & 0.0033 $\pm$ 0.0005 & 0.0601 $\pm$ 0.0054 \\ \hline
Ellipse    & 0.0059 $\pm$ 0.0013 & 0.0025 $\pm$ 0.0003 & 0.0500 $\pm$ 0.0147 \\ \hline
Helix      & 0.0052 $\pm$ 0.0007 & 0.0025 $\pm$ 0.0003 & 0.0422 $\pm$ 0.0093 \\ \hline
\end{tabular}
\label{tab:trajectory_tracking_sim}
\end{table}

In the dynamic obstacle avoidance task, the robot follows a circular trajectory of radius 0.05~m while avoiding collisions with two oscillating obstacles moving in the $xy$-plane at different heights~(see Fig.~\ref{fig:obs}). Unlike tip-based tracking, the controller must adapt the entire body configuration to ensure safe clearance along the backbone. The MPPI cost formulation plays a central role in this behavior. In the \textit{running cost}, body shape deviation from the previous configuration is penalized with a weight of 10, actuator smoothness is enforced with a penalty of 10 on changes relative to the last action, and tip tracking error carries the highest weight of 100 to prioritize trajectory accuracy. Obstacle avoidance is handled as a hard constraint by imposing a large penalty ($10^{5}$) whenever any part of the backbone comes within 0.02~m of either obstacle. The \textit{terminal cost} reinforces these objectives at the prediction horizon. As shown in Fig.~\ref{fig:obs_plot}, the tracking error remains on the order of a few millimeters, with only minor spikes during avoidance maneuvers. The control effort gradually increases over time, reflecting the additional actuation required to both follow the circular path and reshape the body for obstacle clearance. Meanwhile, the minimum distance to obstacles never falls below the 0.02~m threshold, confirming safe operation. These results highlight how the combination of weighted penalties in the running and terminal costs allows the controller to achieve precise trajectory tracking while dynamically adapting its body shape to ensure obstacle avoidance.

\begin{figure}[!t]
    \centering
    \includegraphics[width=0.95\linewidth]{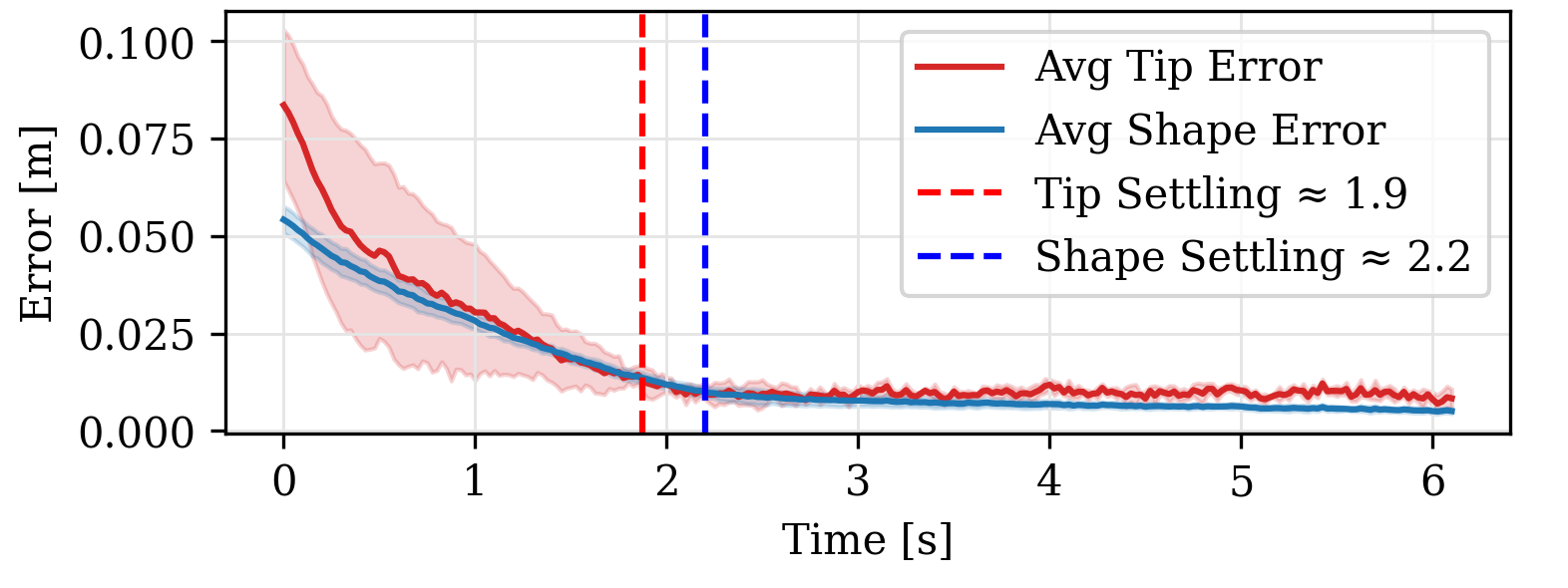}
\caption{Average results of 50 trials for the shape-constrained reaching task. 
The robot achieves a final tip error of 0.0084~m ($\pm$0.0024) and a final shape error of 0.0051~m ($\pm$0.0007). 
Settling times are approximately 1.9~s for the tip and 2.2~s for the shape.}
        \label{fig:shape_res}
\end{figure}

In the shape-constrained reaching task, the robot is required not only to reach a specified target position but also to conform its backbone to a desired reference shape, making this scenario significantly more challenging than tip-based tracking alone (see Fig.~\ref{fig:shape_constrained_reaching}). The results averaged over 50 trials, illustrated in Fig.~\ref{fig:shape_res}, demonstrate that the proposed MPPI-based controller is able to simultaneously satisfy both objectives. The tip converges to the target with a final average error of 0.0084~m ($\pm$0.0024), while the backbone shape deviation from the reference remains low at 0.0051~m ($\pm$0.0007). The settling time analysis shows that the tip reaches steady-state within approximately 1.9~s, while the body shape conforms within 2.2~s. These results indicate that the controller effectively balances end-point accuracy with whole-body coordination, reshaping the continuum structure to follow the specified path while precisely positioning its tip. This highlights the capability of the approach to enforce global shape constraints, which is essential for tasks where both tip positioning and body conformance play a critical role, such as navigation through confined anatomical pathways or interaction with flexible environments.

\begin{figure}[!t]
 \centering
    \subfigure[]{
        \includegraphics[trim={0cm 0cm 8.2cm 0},clip,width=0.44\linewidth]{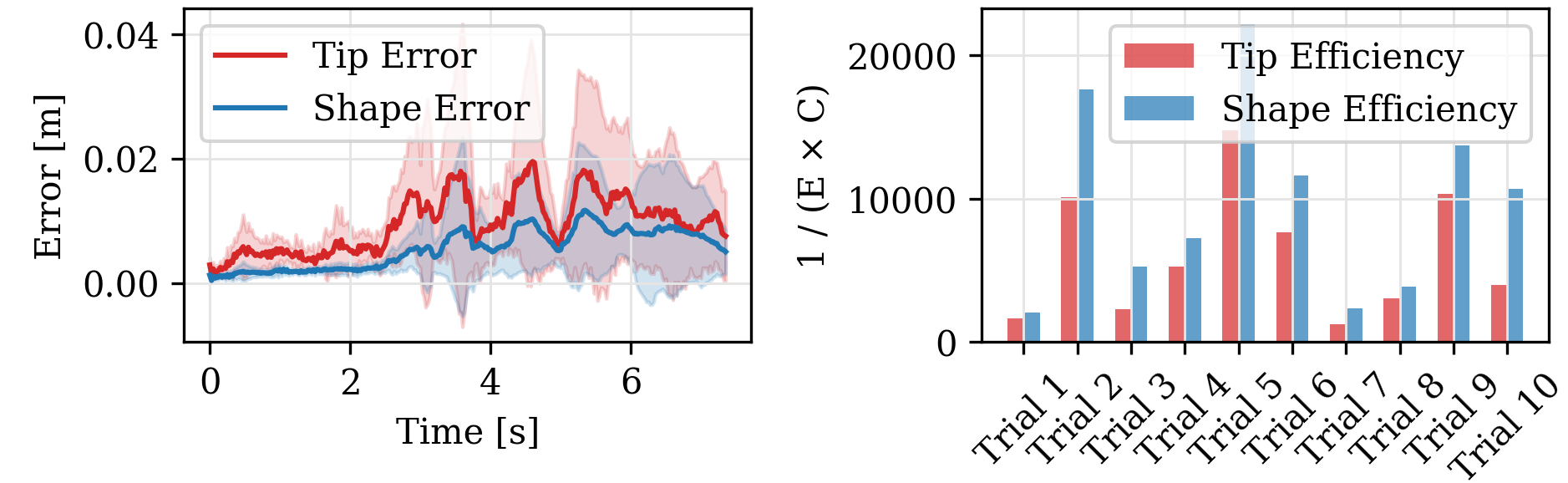}
        \label{fig:teleop_a}
    }
    \subfigure[]{
        \includegraphics[trim={8.cm 0cm 0cm 0},clip,width=0.46\linewidth]{figs/tele_2.png}
        \label{fig:teleop_b}
    }
\caption{Tele-operated shape tracking performance across ten trials. (a)~Average tip tracking error and shape deviation over time, with shaded regions denoting one standard deviation across trials. (b)~Efficiency analysis of each trial, defined as the inverse of the product between final error and command complexity. Taller bars indicate better performance relative to task difficulty, meaning the robot achieved accurate tracking even when the operator’s commands were complex.}
        \label{fig:shape}
        \vspace{-2mm}
\end{figure}

The results of the tele-operated shape tracking task over 10 trials (Fig.~\ref{fig:teleop_a}) demonstrate that the controller is capable of maintaining consistent tracking performance despite large variations in operator command complexity. The average final tip error was approximately $0.0097$~m and the shape deviation $0.0060$~m, indicating that the system achieves millimeter-level accuracy even under tele-operation. To better account for the influence of operator command complexity, we defined an efficiency metric as the inverse of the product between final tracking error and command complexity:
\begin{equation}
    \eta = \frac{1}{E \times C},
\end{equation}
where $E$ denotes the final error (tip or shape) and $C$ quantifies the geometric and temporal complexity of the operator’s commanded shape. This metric highlights how well the robot achieves accurate tracking relative to the difficulty of the task (large $\eta$ means good performance given task difficulty). Figure~\ref{fig:teleop_b} shows that efficiency varies significantly across trials. This variation reflects the fact that operator commands differ in difficulty: simple commands, such as smooth motions, result in high efficiency, while abrupt or highly curved commands reduce efficiency despite similar error magnitudes. Importantly, the most efficient trial was over three times more efficient than the least efficient one, underscoring how the operator’s choice of commands strongly impacts performance. These findings highlight the importance of considering both accuracy and task complexity in tele-operation evaluation, as efficiency better captures how effectively the system executes user-intended motions under varying conditions.

\section{Experiments}
\label{sec:exp}
\begin{figure}[!t]
    \centering
        \includegraphics[width=0.90\linewidth]{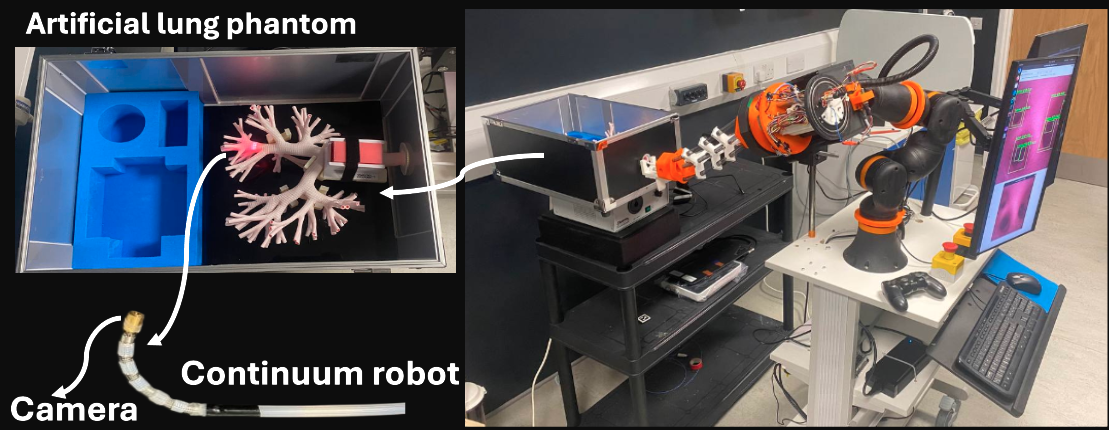}
    \caption{ Experiment setup. A continuum robot with a tip-mounted camera is tele-operated via a joystick to navigate an artificial lung phantom, with visual feedback.}
    \label{fig:setup}
    \vspace{-3mm}
\end{figure}

\begin{figure}[!t]
    \centering
        \includegraphics[width=0.95\linewidth]{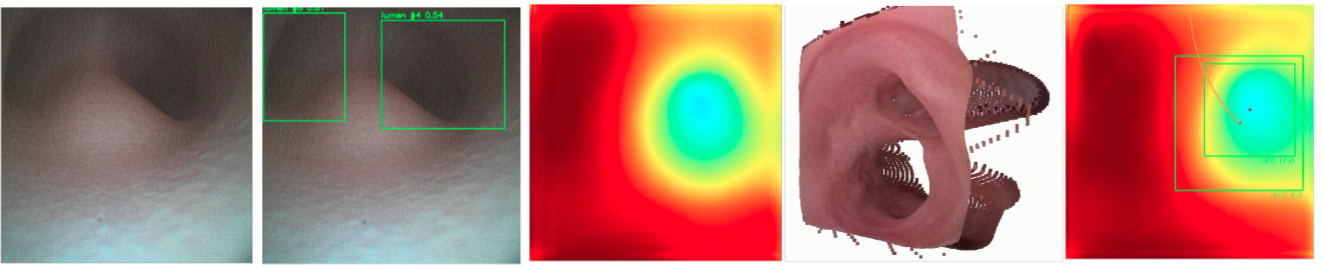}
\caption{Bronchoscopy perception pipeline. 
From left to right: raw bronchoscopy image, lumen detection, depth estimation, generated point cloud of the local airway, and planned trajectory.}
\label{fig:bronchoscopy_pipeline}
    \vspace{-5mm}
\end{figure}

To validate the proposed framework, we employ a bronchoscopy-inspired setup using a tendon-driven continuum robot with a tip-mounted camera, tele-operated via a joystick interface (Fig.~\ref{fig:setup}). The robot is inserted into an artificial lung phantom, and the operator navigates through branching airways based on visual feedback displayed on a monitor. A YOLO-based lumen detector identifies candidate branches, and depth estimation provides local geometry \cite{ZhaFra_BREADepth_MICCAI2025}. Using the depth estimation module, each pixel in the bronchoscopy image is back-projected into 3D via the calibrated camera intrinsics, producing a point cloud of the local airway geometry (Fig.~\ref{fig:bronchoscopy_pipeline}). The operator selects a target lumen, after which a smooth path along the estimated centerline toward the chosen branch is planned. This reference is executed by the shape-aware whole-body controller, while the task manager allows the operator to adjust objectives online, such as emphasizing clearance or direct advancement.

\subsection{Results and Discussions}
We evaluated the framework in a bronchoscopy phantom across 18 subsegmental pulmonary regions. Each region was attempted twice: once using conventional joystick-only teleoperation and once with the proposed framework. Performance was assessed in terms of wall contact frequency, depth of access, and task completion time.  

With joystick-only navigation, operators frequently deviated from the lumen, averaging $7.4 \pm 2.0$ wall contacts per trial, and were able to reach only 12 of the 18 targets (67\%). In contrast, framework-assisted navigation reduced wall contacts to $2.3 \pm 1.1$ per trial and enabled successful access to all 18 targets (100\%), including distal regions that joystick-only control could not consistently reach.  

Task completion times also improved. Framework-assisted trials averaged $54 \pm 6$~s, compared to $61 \pm 8$~s with joystick-only operation, representing a 12\% speed-up ($p<0.05$). The task manager further allowed operators to adapt behavior: prioritizing wall clearance reduced contacts by an additional 30\% at the expense of slightly longer completion times, while prioritizing direct advancement shortened procedures by about 10\% with only a modest increase in contacts.  

Overall, the results show that the proposed framework enables deeper and more reliable access to pulmonary subsegments, while simultaneously reducing wall trauma and improving efficiency compared to joystick-only teleoperation.

\section{Comparison Study}
\label{sec:comp}
We evaluate the proposed framework against end-to-end models~\cite{kasaei2025synergistic}, Neural-ODE-based approaches~\cite{kasaei2023CORL}, and PD-control~\cite{Kasaei2023DataEfficient} in two representative scenarios: butterfly trajectory tracking and shape-constrained reaching. Among these baselines, only~\cite{kasaei2025synergistic} explicitly accounts for backbone shape, although it assumes static obstacles, relies on fixed objectives, and does not include a task manager for online adjustment. The other two approaches are primarily tip-centric and therefore cannot regulate body shape.  

On the butterfly trajectory, all methods demonstrate broadly similar performance, with only marginal differences in tip tracking accuracy. In fact, comparable results can often be achieved across baselines through extensive parameter tuning. Nevertheless, only the shape-aware method of~\cite{kasaei2025synergistic} and our framework explicitly regulate backbone configuration. Unlike~\cite{kasaei2025synergistic}, our controller integrates dynamic obstacle handling and a task manager, enabling online adjustment of objectives such as wall clearance versus direct advancement. This added flexibility provides clear advantages in tele-operated scenarios where task requirements change during execution.  

For the shape-constrained reaching task, we re-trained the method of~\cite{kasaei2025synergistic}, while our controller requires no offline training. To ensure fairness, we extended their method by adding a shape term to the loss function and fine-tuned its parameters for the best performance. With these modifications, their approach achieved much faster settling times and quicker convergence compared to ours. However, this improvement comes at the expense of requiring large amounts of training data and significant parameter tuning. In contrast, our framework achieves comparable accuracy without any training while still supporting online adaptation through the task manager.  

These results underline a key distinction: while existing methods can approach similar levels of tracking accuracy with sufficient tuning and data, our framework uniquely combines shape-aware whole-body control, dynamic obstacle handling, and task-level adaptability without relying on extensive training.

\section{Conclusion}
\label{sec:conclusion}
We have introduced a unified framework for shape-aware whole-body control of continuum robots that integrates physics-informed modeling, sampling-based optimal control, and task-level adaptability. By augmenting Cosserat rod theory with neural residuals, the framework provides accurate backbone estimation while remaining computationally efficient. The MPPI controller exploits this representation to achieve simultaneous tip tracking, shape regulation, and obstacle avoidance, while the task manager enables operators to tune objectives online and adapt robot behavior to changing task demands.  
Through simulations, we demonstrated robust performance across diverse scenarios, including challenging trajectories, dynamic obstacle avoidance, and shape-constrained reaching. Real-world experiments in a bronchoscopy phantom confirmed that our approach improves lumen-following accuracy, reduces wall contact, and increases operator efficiency compared to joystick-only navigation and state-of-the-art baselines.  

Future work will focus on extending the perception and planning modules to handle highly dynamic anatomical environments, improving real-time computational efficiency, and exploring integration with learning-based intent prediction to further enhance shared-control capabilities. 

% The presented framework offers a promising step toward safe, adaptive, and clinically relevant deployment of continuum robots in surgical and other confined environments.

\bibliographystyle{IEEEtran}
\bibliography{main.bbl}  % .bib

\end{document}